\documentclass[journal]{IEEEtran}
\ifCLASSINFOpdf
\else
\fi
\usepackage{times}
\usepackage{epsfig}
\usepackage{graphicx}
\usepackage{amsmath}
\usepackage{amssymb}

\usepackage{subfigure}
\usepackage{amsthm}

\usepackage{multirow}
\usepackage{booktabs}
\usepackage{algorithm}
\usepackage{algorithmic}
\usepackage{epstopdf}
\usepackage{color}
\usepackage{hyperref}

\usepackage{makecell}

\begin{document}
%
\title{Modal-aware Features for Multimodal Hashing}
%
%
%

\author{Haien Zeng, Hanjiang Lai, Hanlu Chu, Yong Tang, Jian Yin
\thanks{Email address: zenghen@mail2.sysu.edu.cn (H. Zeng), (laihanj3, issjyin)@mail.sysu.edu.cn (H. Lai, J. Yin), (hlchu, ytang)@m.scnu.edu.cn (H. Chu, Y. Tang)}
\thanks{H. Zeng, H. Lai, and J. Yin are with School of Data and Computer Science, Sun Yat-Sen University, China. H. Chu and Y. Tang are with School of Computer Science, South China Normal University, China. Hanjiang Lai is Corresponding author.}
}

\markboth{}%
{Shell \MakeLowercase{\textit{et al.}}: Bare Demo of IEEEtran.cls for Journals}
%



\maketitle

\begin{abstract}
Many retrieval applications can benefit from multiple modalities, e.g., text that contains images on Wikipedia, for which how to represent multimodal data is the critical component. Most deep multimodal learning methods typically involve two steps to construct the joint representations: 1) learning of multiple intermediate features, with each intermediate feature corresponding to a modality, using separate and independent deep models; 2) merging the intermediate features into a joint representation using a fusion strategy. However, in the first step, these intermediate features do not have previous knowledge of each other and cannot fully exploit the information contained in the other modalities. In this paper, we present a modal-aware operation as a generic building block to capture the non-linear dependences among the heterogeneous intermediate features  that can learn the underlying correlation structures in other multimodal data as soon as possible. The modal-aware operation consists of a kernel network and an attention network. The kernel network is utilized to learn the non-linear relationships with other modalities. Then, to learn better representations for binary hash codes, we present an attention network that finds the informative regions of these modal-aware features that are favorable for retrieval. Experiments conducted on three public benchmark datasets demonstrate significant improvements in the performance of our method relative to state-of-the-art methods.
\end{abstract}


%
\IEEEpeerreviewmaketitle

\section{Introduction}
Multimodal hashing~\cite{wang2015deep} is a task of embedding multimodal data into a single binary code, which aims to improve performance by using complimentary information provided by the different types of data sources. Since good representations are important for multimodal hashing, in this paper, we focus on developing a better feature learning approach.

To learn the representations, multimodal fusion~\cite{baltruvsaitis2019multimodal} is proposed, which aims to generate a joint representation from two or more modalities in favor of the given task. Multimodal fusion can be mainly divided into two categories~\cite{baltruvsaitis2019multimodal}: model-agnostic approaches~\cite{d2015review} and model-based approaches~\cite{liu2014multiple}. The model-agnostic methods do not use a specific machine learning method. According to the data processing stage, model-agnostic methods  can be mainly split into \textit{early} and \textit{late} fusion. Early fusion immediately combines multiple raw/preprocessed data into a joint representation. In contrast, late fusion performs integration after all of the modalities have made decisions. The model-based approaches fuse the heterogeneous data using different machine learning models, e.g., multiple kernel learning~\cite{gonen2011multiple}, graphical models~\cite{fidler2013sentence} and neural networks~\cite{rajagopalan2016extending}.

\begin{figure}[t]
  \centering
    \includegraphics[width=1\hsize \hspace{0.01\hsize}]{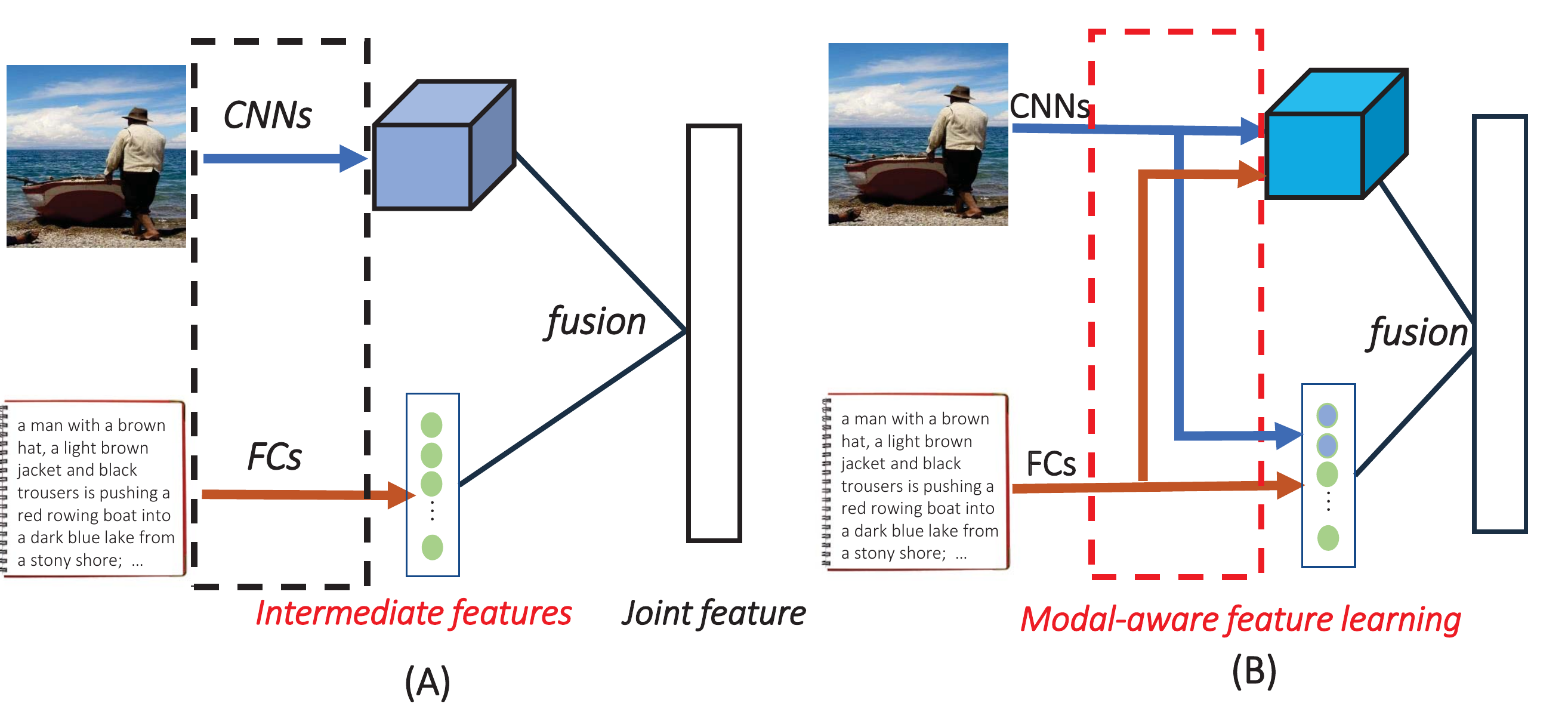}
  \caption{Illustration of two different feature extractions for multimodal data: (A) each modality uses individual neural layers to learn intermediate feature; (B) our proposed modal-aware feature learning that can learn the non-linear dependences among the heterogeneous data. }
  \label{motivation}  
\end{figure}

Recently, deep multimodal fusion has attracted much  attention because it is able to extract powerful feature representations from raw data. As shown in Figure~\ref{motivation} (A), the common practices for deep multimodal fusion are as follows ~\cite{antol2015vqa,mroueh2015deep,ouyang2014multi}: 1) Each modality start with several individual neural layers to learn \textbf{intermediate feature}. 2) These multiple intermediate features are merged into a joint representation via a fusion strategy. Such a fusion approach is referred to as \textit{intermidiate fusion}~\cite{kim2018robust} because the powerful intermediate features obtained by deep neural networks (DNNs) are merged to construct the joint representation. Deep multimodal learning has been shown to achieve remarkable performance for many machine learning tasks, such as deep cross-modal hashing~\cite{DCMH} and deep semantic multimodal hashing~\cite{jin2019deep}.

While they have achieved great success, most existing methods focus on designing better fusion strategies, e.g.,  gated multimodal units (GMUs)~\cite{arevalo2017gated} and multimodal compact bilinear pooling (MCB)~\cite{fukui2016multimodal} for data fusion, and only limited attention has been paid to the intermediate features. The multiple intermediate features are separately learned and do no fully utilize the underlying correlation structures  in other modalities. Thus, a natural question arises: \textit{Can we incorporate the information from other modalities to learn the intermediate features? }

In this paper, we propose a modal-aware operation as a generic building block to learn the multiple intermediate features. Unlike in other deep multimodal approaches, in which each intermediate feature is learned via several individual neural layers, our method learns the dependent and joint intermediate features via the proposed modal-aware operation. The features are forwarded to the modal-aware operation to produce new intermediate features, in which these new intermediate features are learned jointly and dependently and each intermediate feature consists of information from other modalities.

In the context of multimodal hashing, two factors are considered in the proposed modal-aware operation. The first consideration is how to learn the the non-linear dependences from other modalities. Inspired by the kernel methods~\cite{gonen2011multiple}, we present a kernel network to learn the  underlying correlation structures in other modalities. Given two intermediate features from two modalities, we first calculate the kernel similarities, i.e., dot-product similarities, between the two features. Then, the similarities are used as weights to reweight the original features. The second consideration is how to learn better intermediate features for binary hash codes. The binary representations always introduce information loss compared to the original real values, e.g., each bit has only two values: 0 or 1. To reduce the information loss, we further propose an attention network that focuses on selecting the informative parts of multimodal data. The uninformative parts will be removed and will not be used to encode the binary codes. Thus, this method is able to alleviate the information loss to some extent because the binary codes are generated from the informative parts of multimodal data that are favorable for retrieval. To fully utilize the modalities, all of the intermediate features are incorporated to learn the attention maps.

The main contributions of this paper can be summarized as follows.
\begin{itemize}
\item We propose a modal-aware operation to learn the intermediate features. This operation can learn the information contained in other modalities prior to fusion, which is helpful for better capturing data correlations.

\item We propose a kernel network to capture the non-linear dependences and an attention network to find the informative regions. These two networks learn better intermediate features for generating binary hash codes.

\item We conduct extensive experiments on three multimodal databases to evaluate the usefulness of the proposed modal-aware operation. Our method yields better performance compared  with several state-of-the-art baselines.
\end{itemize}

\section{Related Work}

\subsection{Multimodal Fusion} Multimodal fusion is an important step for multimodal learning.  A simple approach for multimodal fusion is to concatenate or sum the features to obtain a joint representation~\cite{kiela2014learning}. For instant, Hu et al.~\cite{hu2016segmentation} concatenated text embeddings and visual features for image segmentation. Reconstruction methods were also proposed to fuse the multimodal data. For example, autoencoders~\cite{ngiam2011multimodal} and deep Boltzmann machines~\cite{srivastava2012multimodal} were trained to reconstruct both modalities with only one modality as the input.
Subsequently, inspired by the success of bilinear pooling and gated recurrent networks, Fukui et al.~\cite{fukui2016multimodal} proposed multimodal compact bilinear pooling to efficiently combine multimodal features, and John et al.~\cite{arevalo2017gated} proposed a gated multimodal unit to determine how much each modality affects unit activation. Liu et al.~\cite{liu2018learn} multiplicatively combined a set of mixed source modalities to capture cross-modal signal correlations.  Although many approaches have been proposed for multimodal fusion, these deep learning methods do not fully explore the dependences among the modalities prior to the fusion operations. In this paper, we argue that capturing the dependences among the heterogeneous modalities will benefit multimodal fusion.

\subsection{Multimodal Retrieval} A similar work is that on cross-modal hashing~\cite{wang2016comprehensive}. Given a query of one modality, the goal of cross-modal hashing is to retrieve the relevant data from another modality. For example, Cross-view hashing (CVH)~\cite{sun2008least} and semantic correlation maximization (SCM)~\cite{zhang2014large} use hand-crafted features. Deep cross modal hashing (DCMH)~\cite{DCMH} and pairwise relationship guided deep hashing (PRDH)~\cite{yang2017pairwise} are deep-network-based methods. Attention-aware deep adversarial hashing~\cite{zhang2018attention} and self-supervised adversarial hashing (SSAH)~\cite{li2018self} apply the adversarial learning to generate better binary codes. Although many approaches have been proposed for cross-modal hashing, our multimodal hashing is different from the cross-modal hashing. The proposed multimodal hashing aims to learn  the \textit{joint representations} but not \textit{coordinated representations}, in which the joint approach combines multiple samples into the same representation space while coordinated approach process the multiple data separately and enforce similarity-preserving among different modalities~\cite{baltruvsaitis2019multimodal}. 

Other similar works include those on multi-view hashing that leverages multiple views to learn better binary codes. Some represetative studies focus on multiple feature hashing (MFH)~\cite{kim2012sequential}, composite hashing with multiple information sources (CHMIS)~\cite{zhang2011composite}, multi-view latent hashing (MVLH)~\cite{shen2015multi}, dynamic multi-view Hashing (DMVH)~\cite{xie2017dynamic} and so on. In this paper, we only consider the multimodal data but not the multiple views, e.g., SIFT and HOG  from the same image modality.

Limited attention has been paid for multimodal hashing. Wang et al.~\cite{wang2015deep} proposed deep multimodal hashing with orthogonal regularization to exploit the intra-modality and inter-modality correlations. Cao et al.~\cite{cao2014medical} proposed an extended probabilistic latent semantic analysis (pLSA) to integrate the visual and textural information. In this paper, we focus on learning better intermediate features for multimodal hashing.

\begin{figure*}[t]
  \centering
    \includegraphics[width=0.95\hsize \hspace{0.01\hsize}]{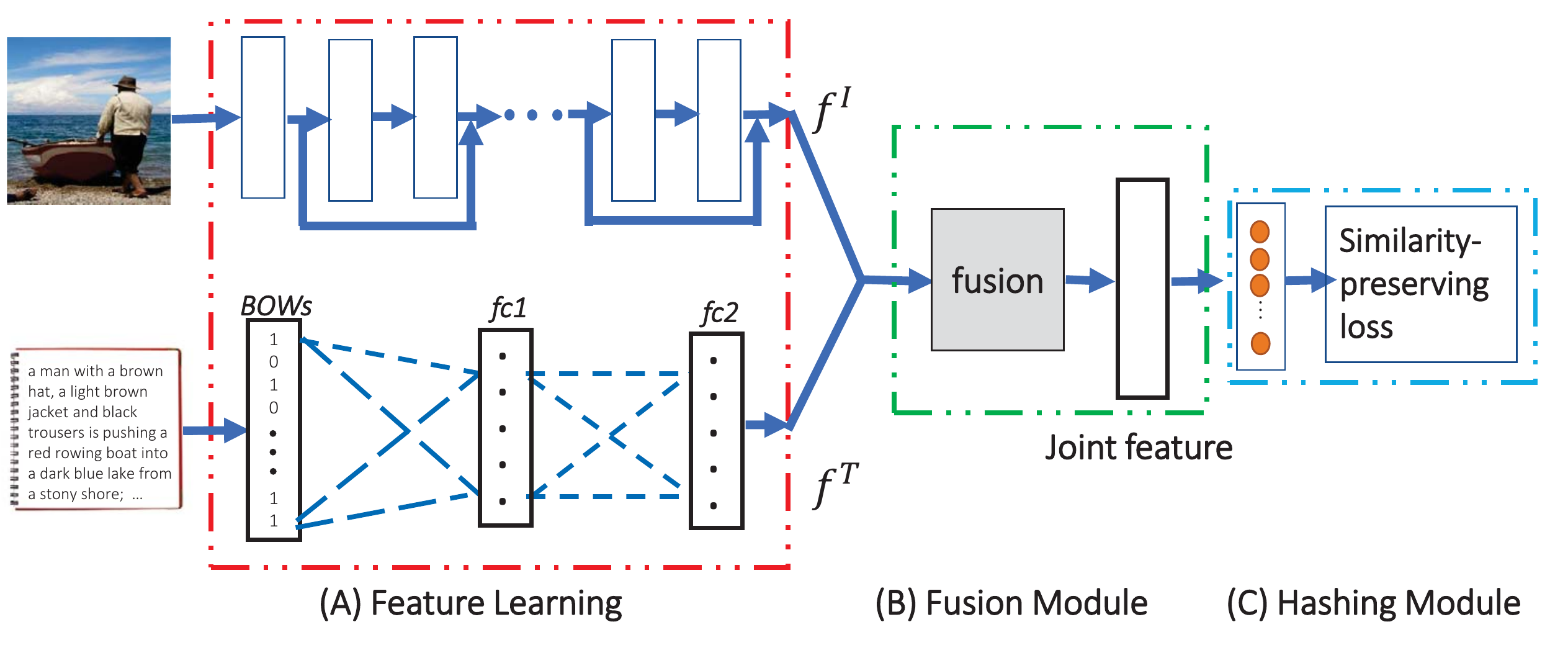}
  \caption{Overview of deep multimodal hashing. It consists of three sequential parts: (A) feature learning module; (B) fusion module; and (C) hashing module. Please note that the intermediate features are learned separately. In this paper, we focus on learning better intermediate features.
  }
  \label{framework}
\end{figure*}

\section{Overview of Deep Multimodal Hashing}

In this section, we briefly summarize the deep multimodal hashing framework.

Let $S = \{S_i \}_{i=1}^n$ denote a set of instances, where each instance is represented in multiple modalities. For ease of presentation, we only consider two modalities, i.e., image and text, to explain our main idea. We denote the instance as $S_i=\{I_i,T_i,Y_i\}$, where $I_i$ and $T_i$ are image and text descriptions of the $i$-th instance, and $Y_i$ is the corresponding ground-truth label. Let $H = \{H_i\}_{i=1}^n$ denote the binary codes, where $H_i \in \{-1,1\}^{l}$ is the $l$-dimensional binary code associated with $S_i$. The aim of multimodal hashing is to learn hash functions that encode the instance $S_i$ into one binary code $H_i$ while preserving the similarities between the instances. For example, if $S_i$ and $S_j$ are similar, the Hamming distance between $H_i$ and $H_j$ should be small. When $S_i$ and $S_j$ are dissimilar, the Hamming distance  should be large.

Different from unimodal data, each instance consists of multiple unimodal signals.  Combining these signals into a joint representation becomes a critical step. Currently, the deep multimodal learning (DML) approaches have been shown to achieve remarkable performance because they can learn the powerful features from all of the modalities. Merging these powerful features into a joint representation will lead to better and flexible multimodal fusion.

An illustration of a deep network for multimodal hashing is shown in Figure~\ref{framework}. The network is divided into three sequential parts: 1) the feature learning module, which learns the efficient intermediate features from the image and text raw data; 2) the multimodal fusion module, which merges the two intermediate features into a joint representation; and 3) the hashing module, which encodes the joint representations to the binary codes, followed by a similarity-preserving loss.

In the feature learning module, the convolutional layers are applied to produce powerful feature maps for the image modality. The images go through several convolutional layers to obtain high-level intermediate feature maps. For the text modality, the feed-forward neural network with stacked fully-connected layers is utilized to encode the text into semantic text features.

In the fusion module, with two intermediate features, a fusion strategy is utilized to obtain a joint representation. Many methods for fusion have been proposed, e.g., concatenation, gate multimodal units (GMUs)~\cite{arevalo2017gated} and multimodal compact bilinear pooling (MCB)~\cite{fukui2016multimodal}.

In the hashing module, the joint representation is mapped into a feature vector with the desired length, e.g., an $l$-bit approximate binary code. Then, the similarity-preserving loss is used to preserve the relative similarities of multimodal data.

However, in the above deep multimodal hashing, these intermediate features are learned separately and had no prior knowledge of other modalities before the fusion. In this paper, we present modal-aware operation that aims to learn better intermediate feature representations. It contains a kernel network that aims to learn the correlations among different modalities and an attention network that finds the informative regions. These two aspects are described in detail in the next section.

\begin{figure}[t]
  \centering
    \includegraphics[width=1\hsize \hspace{0.01\hsize}]{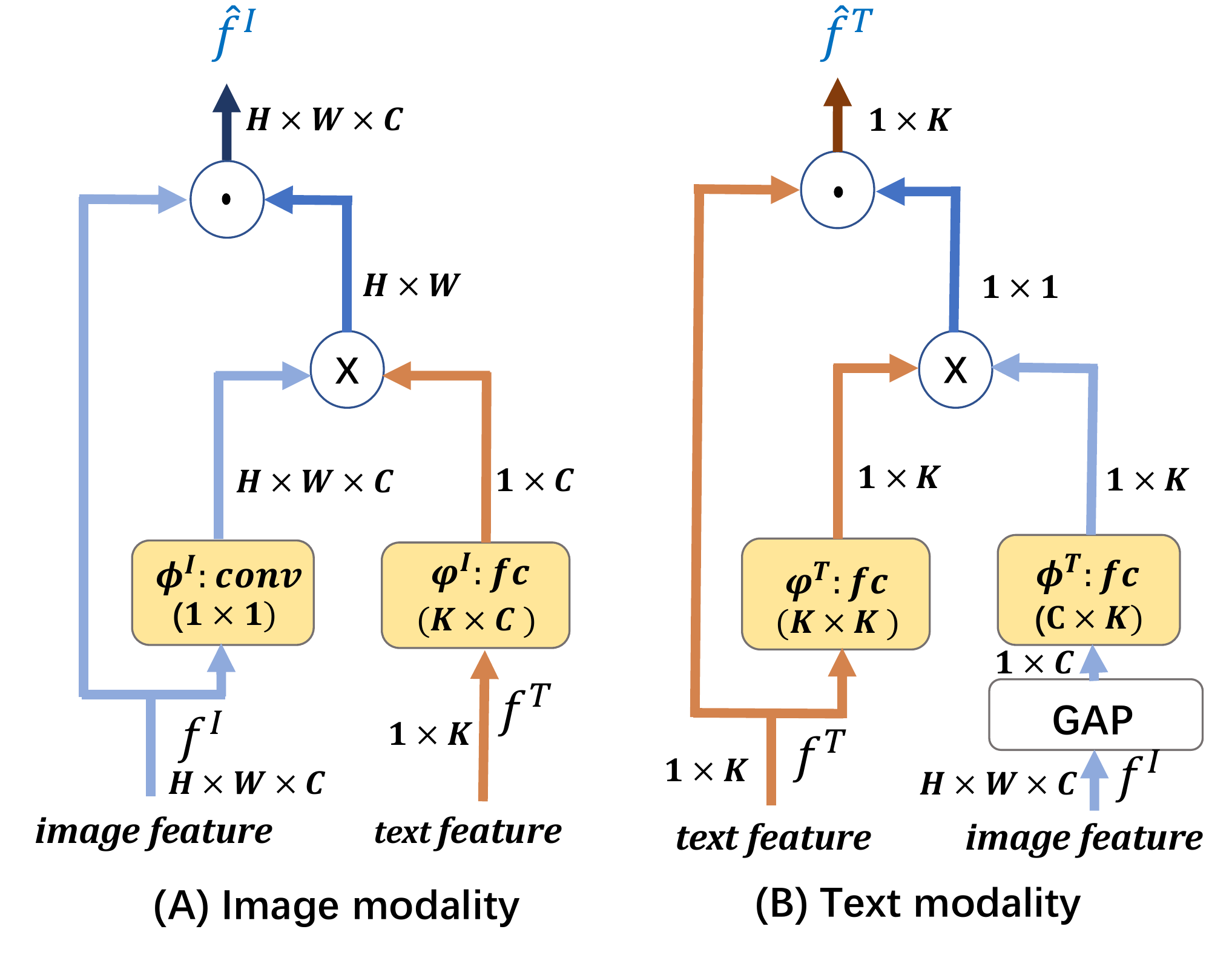}
  \caption{Illustration of the kernel network. The image feature maps $f^I$ with size of $H \times W \times C $ and the text feature vector $f^T$ with feature length $K$. ``$\otimes$" denotes matrix multiplication and ``$\odot$" denotes element-wise multiplication. ``conv" and ``fc" denote the convolutional and fully connected layers, respectively. ``GAP" represents the global average pooling layer.}
  \label{kernel}
\end{figure}

\section{Modal-aware Operation}
In this section, we present a  modal-aware operation that consists of two parts: a kernel network and an attention network.

\subsection{Kernel Network}
The kernel network takes two intermediate features as inputs: image feature maps and the text feature vector. More specifically, suppose that $f^I \in \mathbb{R}^{H \times W \times C}$ represents feature maps for the image modality, where $H, W$ and $C$ are the height, weight and channel, respectively. $f^T \in \mathbb{R}^K$ is the corresponding textural feature, where $K$ is the feature length.

Inspired by the non-local features~\cite{wang2018non} and kernel methods, the outputs of the kernel network are defined as
\begin{equation}
\begin{aligned}
&\hat{f}^I = \mathbf{K}^I(f^I, f^T) f^I,& \\
&\hat{f}^T = \mathbf{K}^T(f^I, f^T) f^T,&
\end{aligned}
\label{imgfeature}
\end{equation}
where $\mathbf{K}^I(x,y)$ and $\mathbf{K}^T(x,y)$ are the kernel functions that measure the similarity between the inputs $x$ and $y$. We use the kernel methods to exploit the correlation structures obtained in other modalities. In Eq.~(\ref{imgfeature}), the intermediate features of the image modality are learned from both the textural and image features. First, the kernel similarity between the image feature and textural feature is calculated. Then, this similarity is used to reweight the original feature. Thus, using these operations, the image feature is embedded into textural information.
The same approach is used for the text modality. We note that we use different kernel functions because the textural feature is a one-dimensional vector while the image feature maps are three-dimensional tensors.

To train the kernel network in an end-to-end manner, the kernel function $\mathbf{K}(x,y)$ is further expressed as the inner product in another space $\mathcal{H}$, which is reformulated as
\begin{equation}
\mathbf{K}(x,y) = \langle \phi(x), \varphi(y) \rangle_{\mathcal{H}},
\end{equation}
where $\phi(\cdot)$ and $\varphi(\cdot)$ are two mapping functions to project the data into another space. Since we use deep networks to learn the multimodal data, we also design two networks as these two mapping functions. That is, the convolutional layer and the fully connected layer are utilized as the mapping functions: $\phi(\cdot)$ is a convolutional layer and $\varphi(\cdot)$ is a fully-connected layer. 

Figure~\ref{kernel} shows the specific structure of the kernel network. For the image modality, the network takes the feature maps $f^I$ and the textural vector $f^T$ as inputs. The approach consists of three parts: 1) two mapping functions $\phi^I(f^I)$ (a convolutional layer) and $\varphi^I(f^T)$ (a fully connected layer) are first learned; 2) the kernel similarity is calculated using the inner product layer; and 3) the origin features are reweighted using the kernel similarity.  In the first part, $\phi^I$ is a convolutional layer with $1 \times 1$ kernel size, and $\varphi^I$ is a single-layer neural network with transformation matrix $W \in \mathbb{R}^{K \times C}$ that maps the textural feature and the visual features to the same dimension, which is given by
\begin{equation}
V^I = \phi^I(f^I), T^I = \varphi^I(f^T).
\end{equation}
Since $V^I$ is a tensor while $T^I$ is vector, we first reshape the feature maps by flattening the height and width of the original features: $V^I = [V^I_{1},\cdots, V^I_{M}]$, where $V^I_{i} \in \mathbb{R}^C$ and $M = H \times W$. The inner products between these $M$ features and the text feature $T^I$ can be calculated. The output of $\hat{f}^I$ can be defined as
\begin{equation}
\hat{f}^I_i = \langle V^I_i, T^I \rangle \ f^I_i, \ \ \forall i=1,\cdots,M,
\end{equation}
where $\hat{f}^I_i$ is the $i$-th vector corresponding to $V_i^I$.

A similar approach is used for the text modality. First, the global average pooling (GAP) layer reduces  $f^I$ with dimensions $H \times W \times C$ to dimensions $1 \times 1 \times C$ by taking the average of each $H \times W$ feature map. Let $\bar{f}^I$ denote the output vector of the GAP layer. Since $\bar{f}^I$ is a vector, $\phi^T$ and $\varphi^T$ are two fully connected layers:
\begin{equation}
V^T = \phi^T(\bar{f}^I), \ T^T = \varphi^T(f^T),
\end{equation}
where $\phi^T$ is connected with the transformation matrix $W_{\phi^T} \in \mathbb{R}^{C \times K}$ and $\varphi^T$ is connected with the transformation matrix $W_{\varphi^T} \in \mathbb{R}^{K \times K}$. Finally, the output for the text modality can be formulated as
\begin{equation}
\hat{f}^T = \langle V^T, T^T \rangle \ f^T.
\end{equation}

\begin{figure}[t]
  \centering
    \includegraphics[width=1\hsize \hspace{0.01\hsize}]{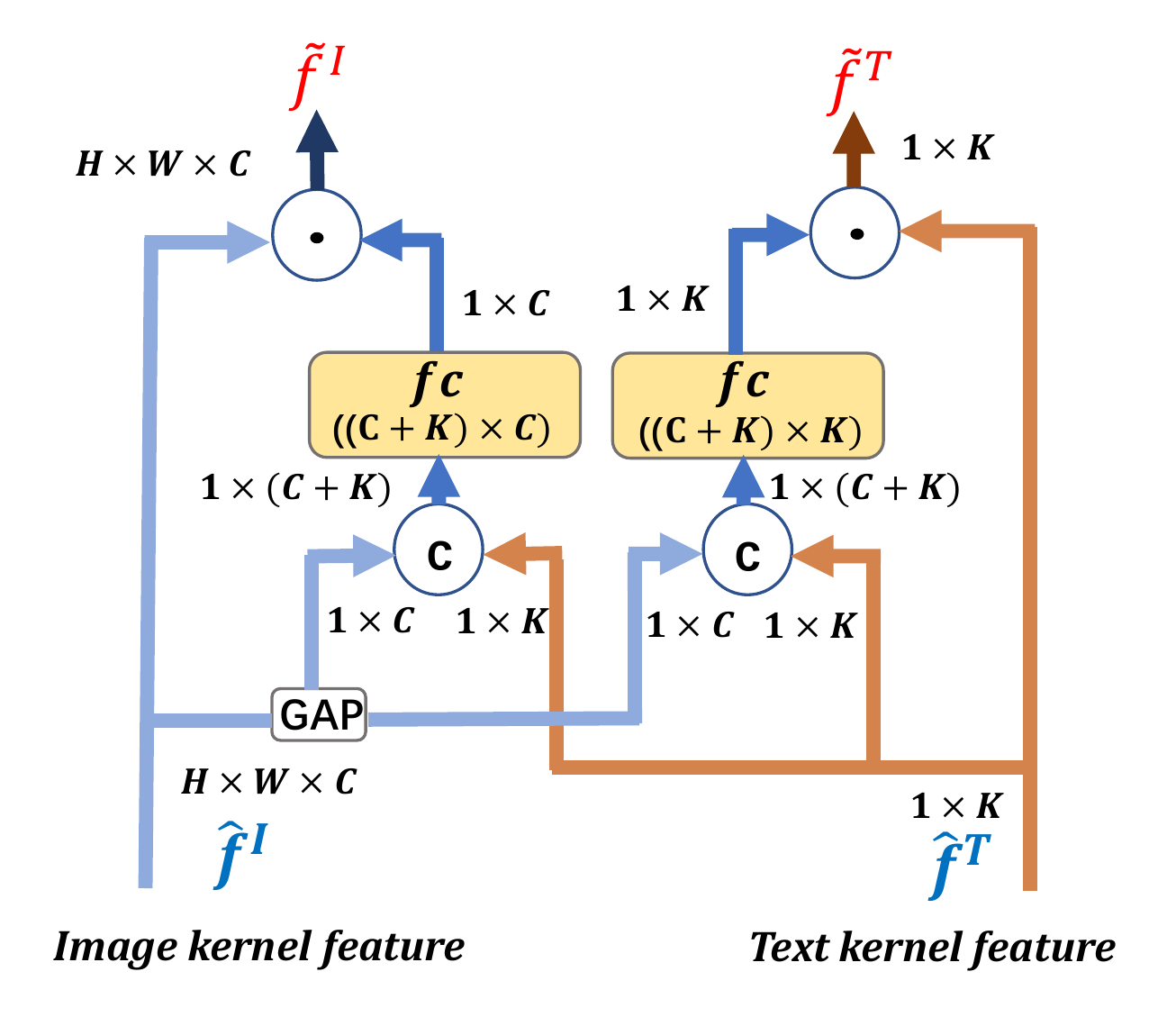}
  \caption{Illustration of the attention network. ``GAP" represents the global average pooling layer, and ``fc" denotes the  fully connected layer. ``$C$" denotes concatenation of two vectors and ``$\odot$" denotes element-wise multiplication.  }
  \label{attention}
\end{figure}

\subsection{Attention Network}
Inspired by how humans process information, we propose an attention network that adaptively focuses on salient parts to learn more powerful multiple intermediate features. To compute the attention efficiently, we aggregate information from all intermediate features. That is, we exploit both features rather than using each independently to locate the informative regions. The detailed operations are described below.

\begin{figure*}[t]
  \centering
    \includegraphics[width=1\hsize \hspace{0.01\hsize}]{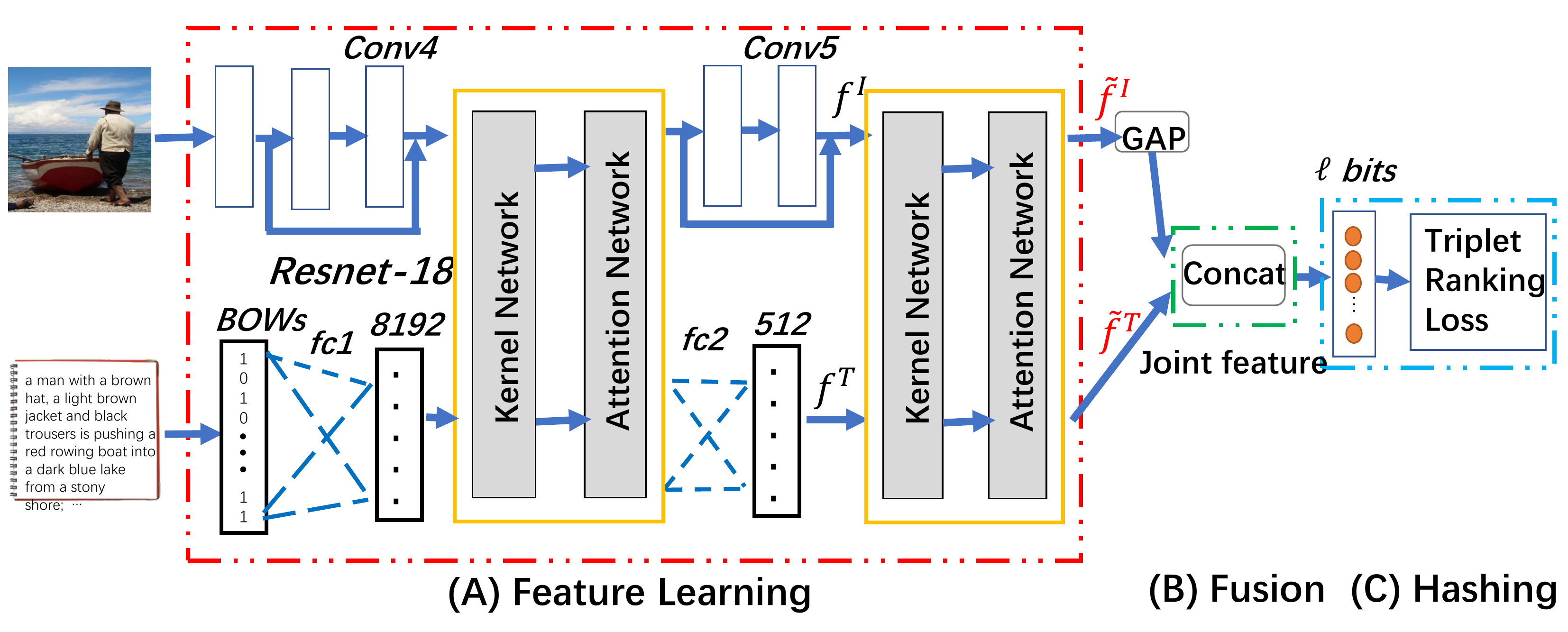}
  \caption{The proposed modal-aware feature learning for multimodal hashing. Two modal-aware operations were added in the feature learning module.  }
  \label{our_framework}
\end{figure*}

Figure~\ref{attention} shows the specific structure of the attention network. First, the visual feature maps $\hat{f}^I$ are forwarded to a global average pooling layer to produce a visual vector $F^I$. Then, we concatenate visual and textural features as $F = [F^I; \hat{f}^T]$, which contains information from different modalities. The feature $F$ goes through two different networks to separately produce attention maps for the image and textural features. Both the networks are composed of a single-layer neural network followed by a softmax function to obtain the attention distributions.
\begin{equation}
\begin{aligned}
& a^I = softmax (W_I F + b_I), & \\
& a^T = softmax (W_T F + b_T), &
\end{aligned}
\end{equation}
where $W_I \in \mathbb{R}^{(C+K) \times C}$ and $W_T \in \mathbb{R}^{(C+K) \times K}$ are transformation matrices. $b_I$ and $b_T$ are model biases. Here, $a^I$ is also called the channel attention map~\cite{woo2018cbam}, which exploits the inter-channel relationship of the features. The main different is that our method use both the visual and textural features from different modalities to find the salient channels. Then, element-wise multiplication is applied to obtain the final outputs $\widetilde{f}^I$ and $\widetilde{f}^T$, which are defined as
\begin{equation}
\begin{aligned}
&\widetilde{f}^I(:,:,i) = a^I_i \hat{f}^I(:,:,i), \ \forall i=1,\cdots,C, & \\
&\widetilde{f}^T_i = a^T_i \hat{f}^T_i, \ \forall i=1,\cdots,K,&
\end{aligned}
\end{equation}
where $\hat{f}(:,:,i)$ is the $i$-th channel with size $H \times W$ and $a_i$ is the $i$-th value in vector $a$.

\section{Implementation Details}
The proposed modal-aware feature learning for multimodal hashing is shown in Figure~\ref{our_framework}. We apply modal-aware operations in the earlier layers. Please note that it only has two fully-connected layers for text modality. Hence, the two modal-aware operations were applied after each fully-connected layer.


\subsection{Network Architectures} For the image modality, ResNet-18~\cite{he2016deep} is used as the basic architecture to learn the powerful image features. ResNet is a residual learning framework that has shown great success in many machine learning tasks. In the ResNet-18, the last global average pooling layer and a 1000-way fully connected layer are removed. The feature maps in Conv4\_2 and Conv5\_2 are used as the image intermediate features $f^I$, respectively. For the text modality, the well-known bag-of-words (BoW) vectors are used as the inputs. Then, the vectors go through a feed-forward neural network  (BoW $\to 8192 \to 512 $) to learn the semantic text features $f^T$.

After the modal-aware operation, we have two features: $\widetilde{f}^I$ and $\widetilde{f}^T$. Since $\widetilde{f}^I$ is a tensor, the global average pooling layer is used to map $\widetilde{f}^I$ into a vector $\widetilde{F}^I$. Then, a simple approach that concatenates these two features is applied to obtain a joint representation. Let $F = [\widetilde{F}^I; \widetilde{f}^T]$ represent the joint representation. The joint representation is forwarded to an $l$-way fully connected layer to generate $l$-bit binary codes $H$.

\subsection{Training Object} We use the triplet ranking loss~\cite{lai2015simultaneous} to train the deep network. We note that other losses, e.g., contrastive loss~\cite{hadsell2006dimensionality}, can also be used in our framework and the loss function is not our focus in this paper. Specifically, given a triplet of instances $(S_i, S_j, S_k)$, in which the instant $S_i$ is more similar to $S_j$ than to $S_k$, these three instances go through the deep multimodal network, and the outputs of the network are $H_i, H_j$ and $H_k$, which are respectively associated with the instances. The similarity-preserving loss function is defined by
\begin{equation}
\begin{aligned}
& \sum_{\langle i, j, k \rangle} \max\{0, \varepsilon + ||H_i - H_j|| - ||H_i - H_k||\}  \\
\end{aligned}
\end{equation}
where $\langle i, j, k \rangle$ is the triplet form and $\varepsilon$ is the margin.

\begin{table*}[t]
\centering
\caption{Comparison with state-of-the-art methods on three datasets.}
\label{table1}
\begin{tabular}{|c|p{0.9cm}p{0.9cm}p{0.9cm}p{0.9cm}|p{0.9cm}p{0.9cm}p{0.9cm}p{0.9cm}|p{0.9cm}p{0.9cm}p{0.9cm}p{0.9cm}|}
\hline
\multirow{2}{*}{{Method}}&\multicolumn{4}{c|}{NUS-WIDE}&\multicolumn{4}{c|}{MIR-Flickr 25k}&\multicolumn{4}{c|}{IAPR TC-12}\\
\cline{2-13}
&16bits & 32bits & 48bits & 64bits & 16bits & 32bits & 48bits & 64bits & 16bits & 32bits & 48bits & 64bits \\
\hline
DPSH &0.7057 &0.7216 &0.7252 &0.7298 &0.8262 &0.8316 &0.8304 &0.8301 & 0.5386 & 0.5448 & 0.5383 & 0.5355\\
\hline
DSH &0.5712 &0.5952 &0.5998 &0.6039 &0.7234 &0.7312 &0.7390 &0.7403 & 0.4746 & 0.4851 & 0.4892 & 0.4926\\
\hline
HashNet &0.7115 &0.7252 &0.7286 &0.7317 &0.8297 &0.8333 &0.8331 &0.8328 & 0.5391 & 0.5451 & 0.5379 & 0.5386\\
\hline
DTH &0.7096 &0.7193 &0.7267 &0.7362 &0.8251 &0.8332 &0.8418 &0.8406 & 0.5662 & 0.5854 & 0.5920 & 0.6032\\
\hline
TextHash &0.6027 &0.6037 &0.6088 &0.6104 &0.7154 &0.7142 &0.7121 &0.7065 & 0.5238 & 0.5487 & 0.5542 & 0.5623 \\
\hline
Concat &0.7274 &0.7391 &0.7432 &0.7495 &0.8352 &0.8453 &0.8554 &0.8508 & 0.5762 & 0.5993 & 0.6213 & 0.6206\\
\hline
GMU &0.7250 &0.7416 &0.7458 &0.7569 &0.8398 &0.8465 &0.8505 &0.8552 & 0.5694 & 0.6006 & 0.6207 & 0.6241\\
\hline
MCB &0.7262 &0.7421 &0.7481 &0.7510 &0.8379 &0.8444 &0.8524 &0.8528 & 0.5721 & 0.5975 & 0.6149 & 0.6151\\
\hline
Ours &{\bfseries 0.7395} &{\bfseries 0.7563} &{\bfseries 0.7627} &{\bfseries 0.7639} &{\bfseries 0.8564} &{\bfseries 0.8658} &{\bfseries 0.8697} &{\bfseries 0.8723} &{\bfseries 0.5925} &{\bfseries 0.6194} &{\bfseries 0.6330} &{\bfseries 0.6384}  \\
\hline
\end{tabular}
\label{table_MAP}
\end{table*}

\section{Experiments}
In this section, we conduct extensive evaluations of the proposed method and compare it with several state-of-the-art algorithms.

\subsection{Datasets}

\begin{itemize}
\item \textbf{NUS-WIDE}~\cite{NUS-WIDE}: This dataset consists of 269,648 images and the associated tags from Flickr. Each image is associated with several textural tags. The text for each point is represented as a 1,000-dimensional bag-of-words vector.

\item \textbf{MIR-Flickr 25k}~\cite{MIR-Flickr}: This dataset contains 25,000 images collected from Flickr. Each image has associated textural tags. The textural tags are represented as a 1,386-dimensional bag-of-words vector.

\item \textbf{IAPR TC-12} ~\cite{IAPR-TC12}: This dataset consists of 20,000 still natural images. Each image is associated with a text caption, which is represented as 2,912-dimensional bag-of-words vector.
\end{itemize}

For all of the experiments, we follow the experimental protocols of DCMH~\cite{DCMH} to construct the query sets, retrieval databases and training sets. The NUS-WIDE dataset contains 81 ground-truth concepts. To prune the data without sufficient tag information, a subset of 195,834 image-text pairs that belong to the 21 most-frequent concepts are selected, as suggested by~\cite{DCMH}. The randomly sampled 2,100 image-text pairs (100 pairs per concept) are used as the query set, and the rest of the image-text pairs are constructed as the retrieval database. In the retrieval database, 10,000 image-text pairs are randomly selected to train the hash functions. In the  MIR-Flickr 25k and IAPR TC-12 databases, the randomly sampled 2,000 image-text pairs are used as the query set. The rest of the pairs are used as the database for retrieval. We randomly select 10,000 pairs from the retrieval database to form the training set.

\subsection{Experimental Settings}
We implement our codes based on the open source deep learning platform PyTorch~\footnote{https://pytorch.org/}.  For the image modality, ResNet-18 is adapted as the basic architecture. The weights of ResNet-18 are initialized with the pretrained model that learns from the ImageNet dataset. For the text modality, the weights of all fully connected layers are randomly initialized following a Gaussian distribution with a standard deviation of 0.01 and a mean of 0. We train the networks by the stochastic gradient solver, i.e., ADAM (weight\_decay = $0.00001$). The batch size is 100, and the base learning rate is 0.0001, which is changed to one-tenth of the current value after every 20 epochs. For fair comparison, all of deep learning methods are based on the same network architectures and same experimental settings.

\begin{figure*}[t]
\subfigure[NUS-WIDE]{
\begin{minipage}[h]{0.33\linewidth}
\centering
\includegraphics[width=1.\textwidth]{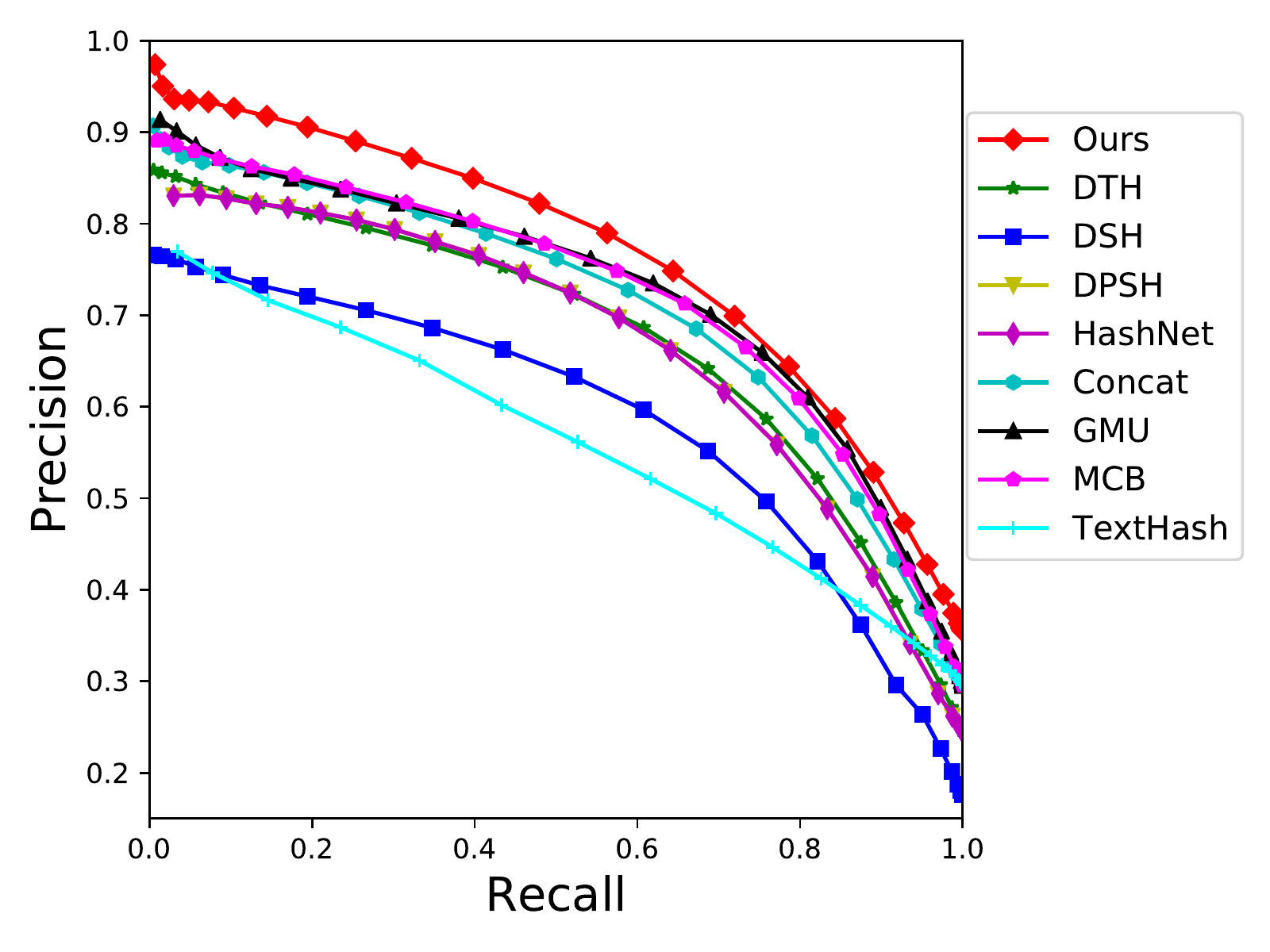}
\label{cifar:side:a}
\end{minipage}%
}
\subfigure[MIR-Flickr 25k]{
\begin{minipage}[h]{0.33\linewidth}
\centering
\includegraphics[width=1.\textwidth]{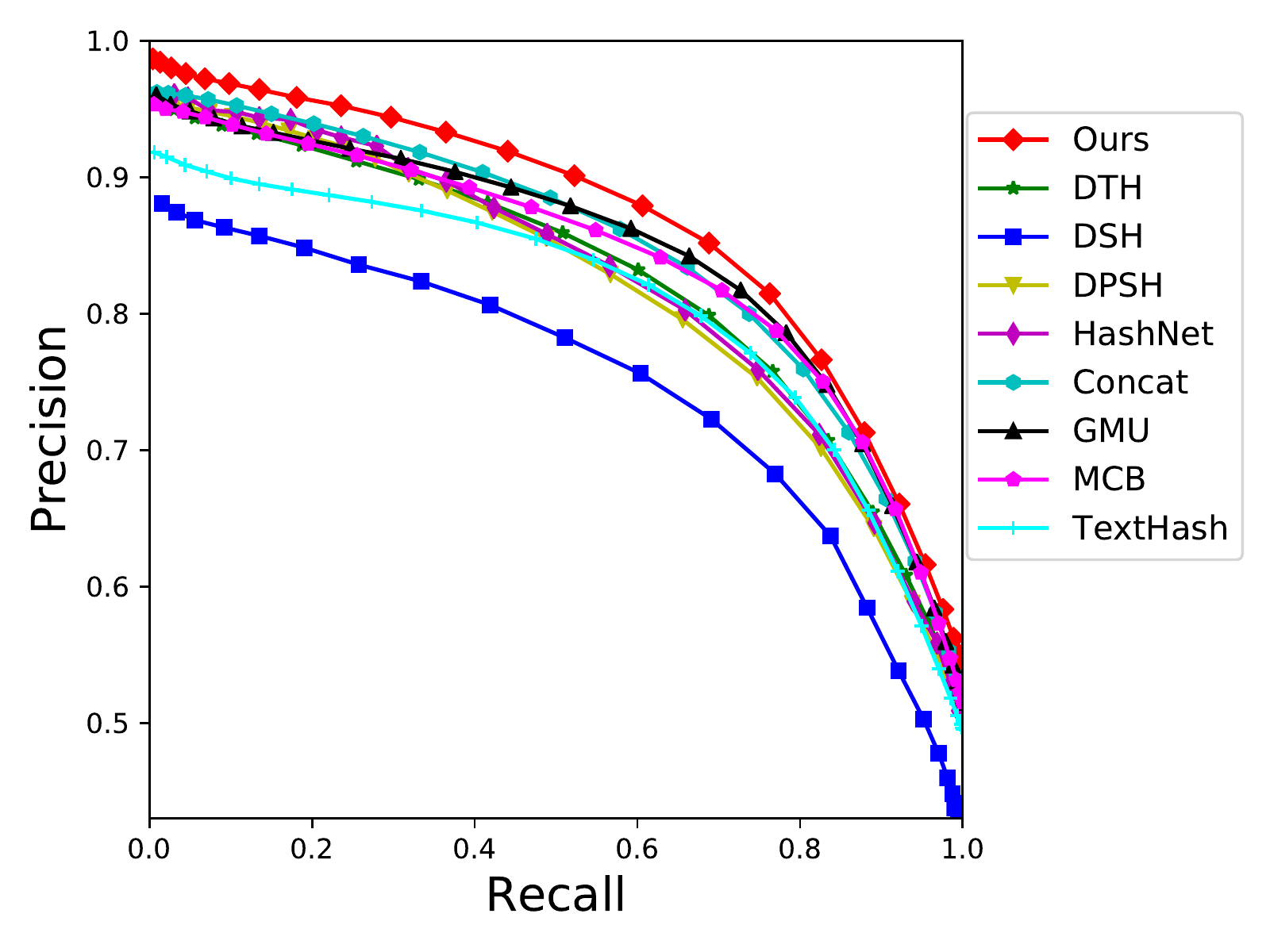}
\label{sun:side:b}
\end{minipage}
}
\subfigure[IAPR TC-12]{
\begin{minipage}[h]{0.33\linewidth}
\centering
\includegraphics[width=1.\textwidth]{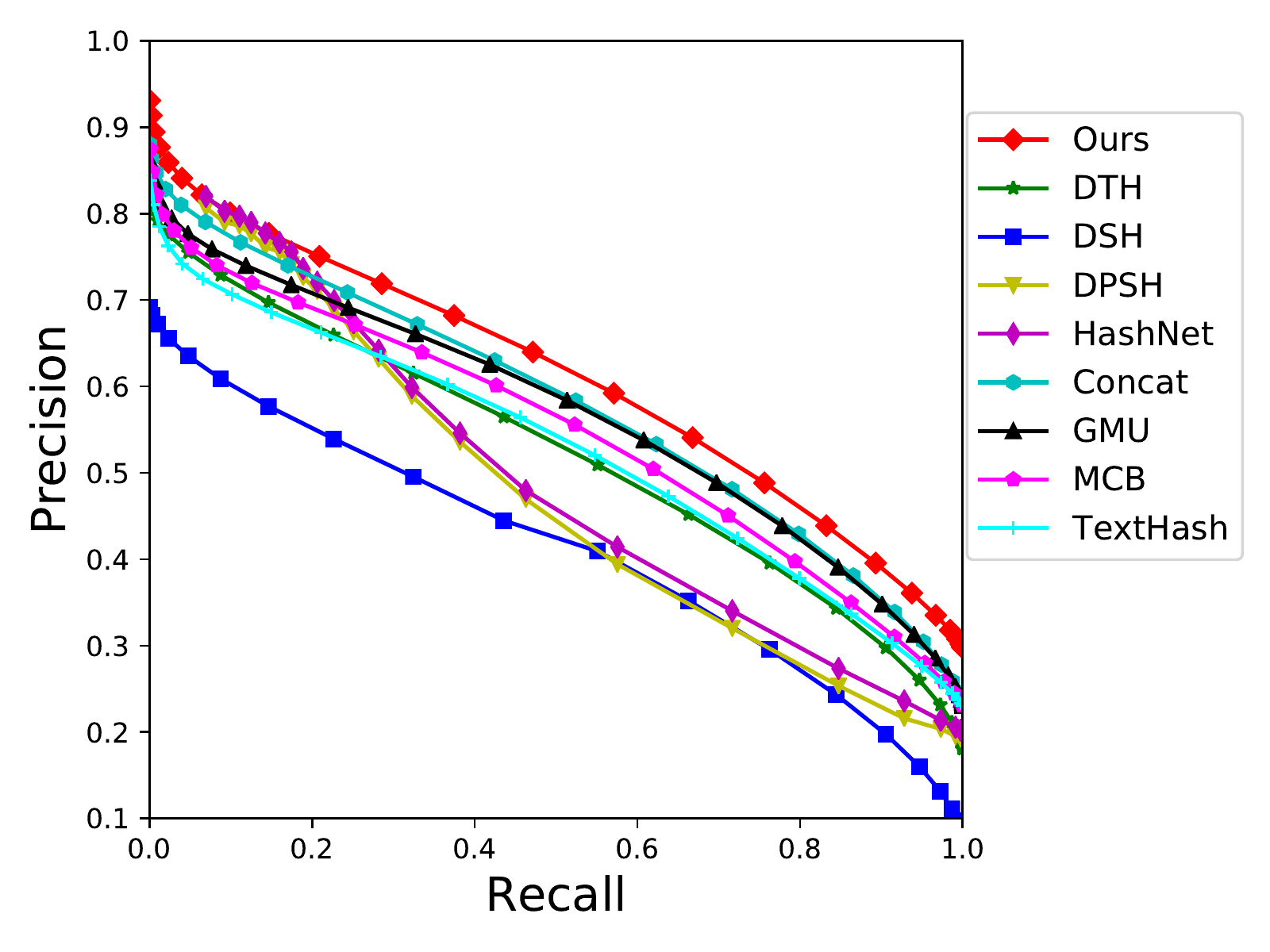}
\label{sun:side:b}
\end{minipage}
}
\caption {\small{The comparison results of precision-recall curves with 32 bits.}}
\label{Precision-recall}
\end{figure*}

\begin{figure*}[t]
\subfigure[NUS-WIDE]{
\begin{minipage}[h]{0.33\linewidth}
\centering
\includegraphics[width=1.\textwidth]{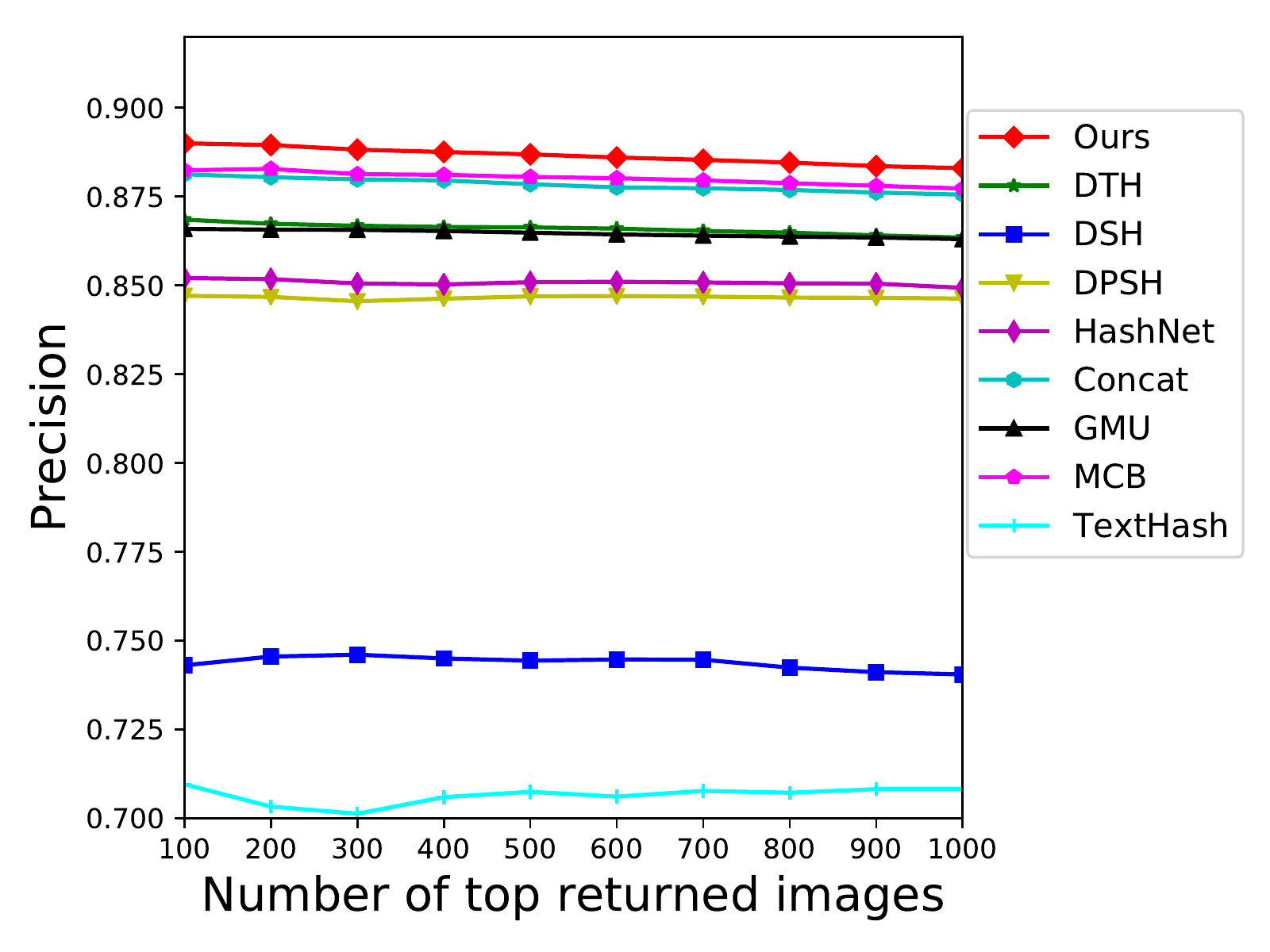}
\label{cifar:side:a}
\end{minipage}%
}
\subfigure[MIR-Flickr 25k]{
\begin{minipage}[h]{0.33\linewidth}
\centering
\includegraphics[width=1.\textwidth]{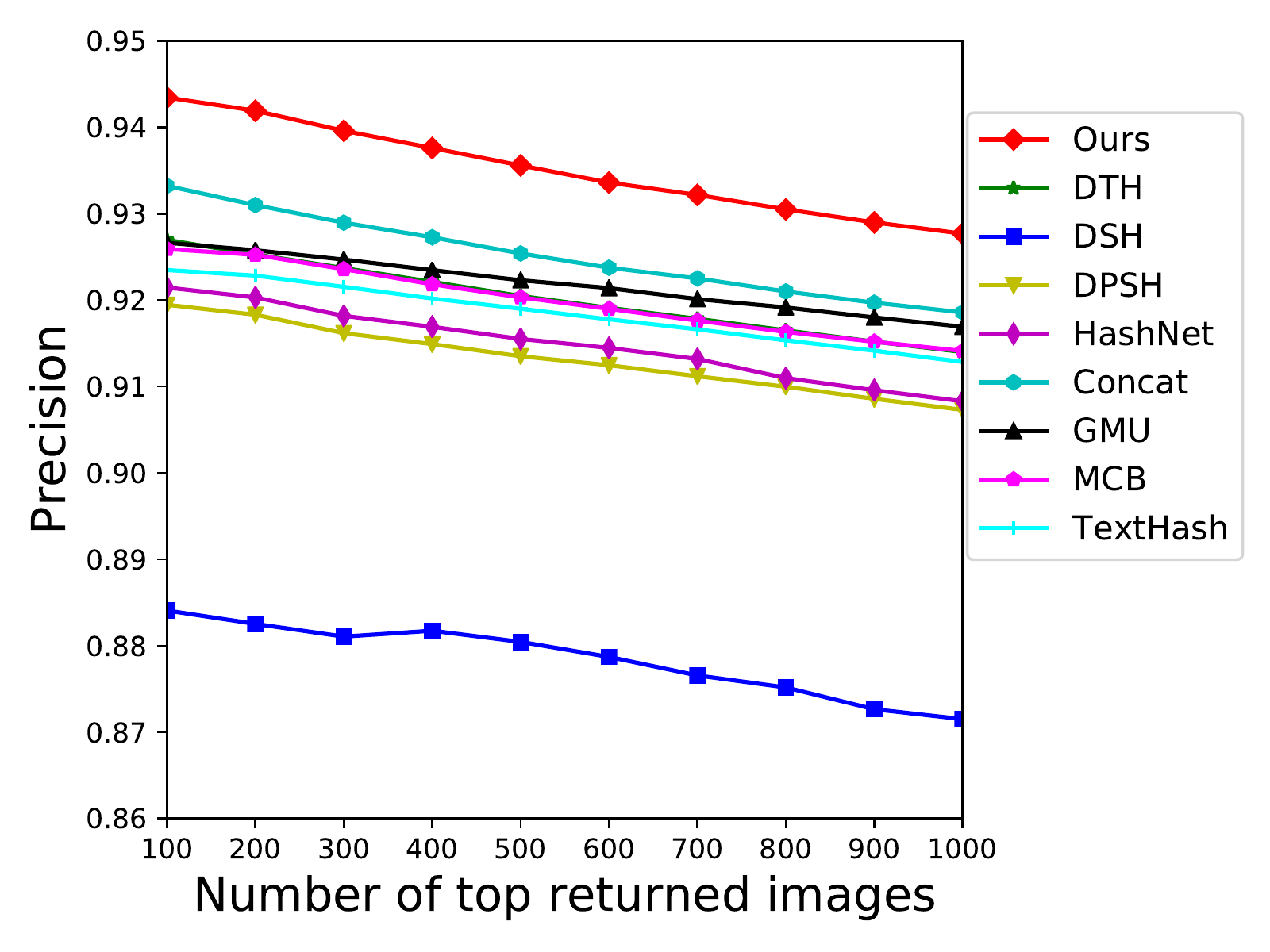}
\label{sun:side:b}
\end{minipage}
}
\subfigure[IAPR TC-12]{
\begin{minipage}[h]{0.33\linewidth}
\centering
\includegraphics[width=1.\textwidth]{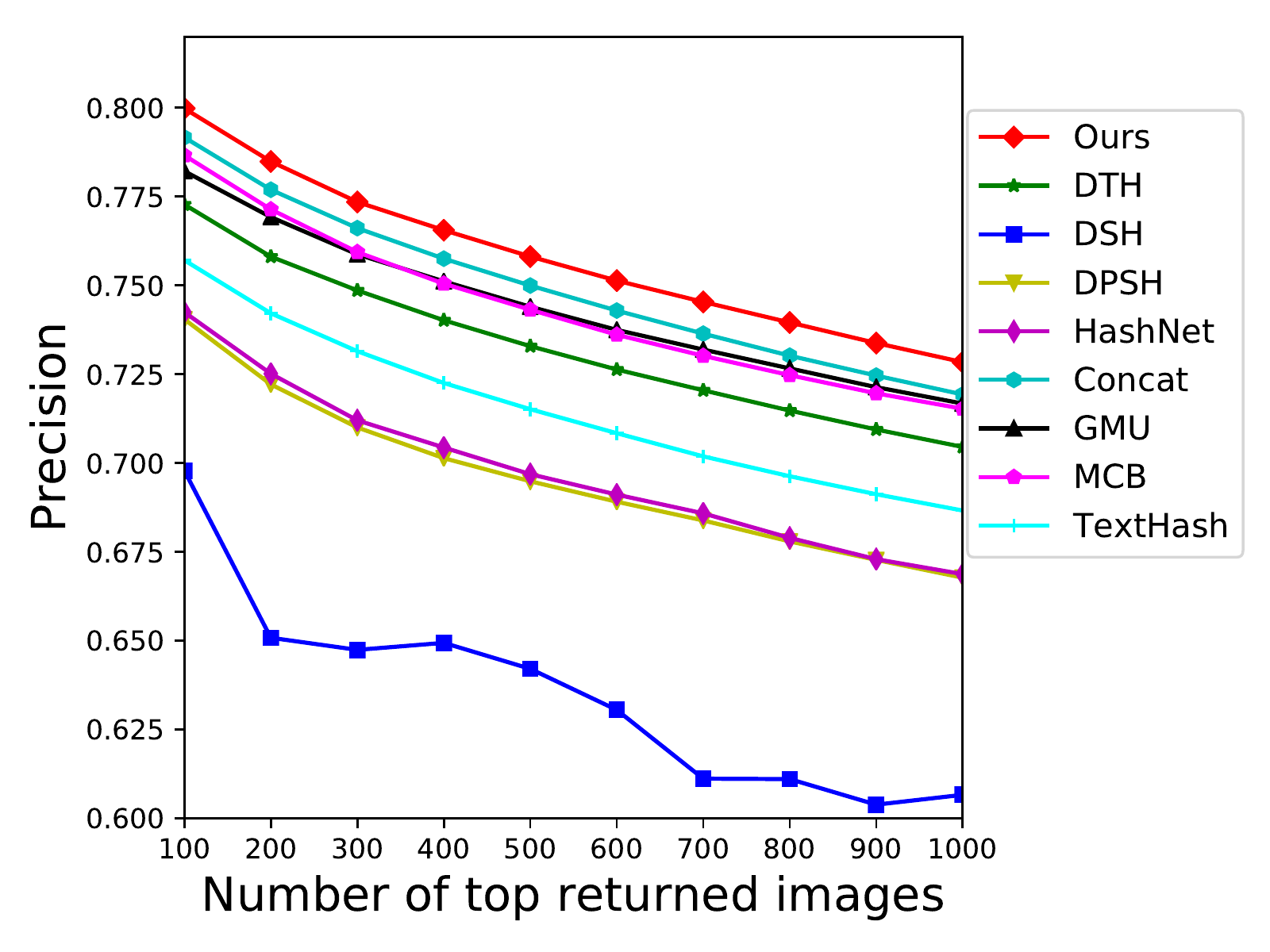}
\label{sun:side:b}
\end{minipage}
}
\caption {\small{The comparison results of precision curve w.r.t. different numbers of top returned samples.}}
\label{Precision}
\end{figure*}

\textbf{Evaluations:} Following the common practice, the mean average precision (MAP), precision-recall and precision w.r.t different numbers of top returned samples are used as the evaluation metrics. MAP is used to measure the accuracy of the whole binary codes based on the Hamming distances. The precision-recall aims to measure the hash lookup protocol and the precision considers only the top returned samples.

\begin{table*}[t]
\centering
\caption{Ablation study on each component on three datasets.}
\label{table1}
\begin{tabular}{|c|p{0.9cm}p{0.9cm}p{0.9cm}p{0.9cm}|p{0.9cm}p{0.9cm}p{0.9cm}p{0.9cm}|p{0.9cm}p{0.9cm}p{0.9cm}p{0.9cm}|}
\hline
\multirow{2}{*}{{Method}}&\multicolumn{4}{c|}{NUS-WIDE}&\multicolumn{4}{c|}{MIR-Flickr 25k}&\multicolumn{4}{c|}{IAPR TC-12}\\
\cline{2-13}
&16bits & 32bits & 48bits & 64bits & 16bits & 32bits & 48bits & 64bits & 16bits & 32bits & 48bits & 64bits \\
\hline
w/o KN &0.7295 &0.7398 &0.7467 &0.7508 &0.8349 &0.8481 &0.8564 &0.8557 & 0.5796 & 0.6037 & 0.6228 & 0.6261\\
\hline
w/o AN &0.7339 &0.7420 &0.7519 &0.7583 &0.8430 &0.8555 &0.8625 &0.8644 & 0.5839 & 0.6073 & 0.6242 & 0.6326\\
\hline
Ours &{\bfseries 0.7395} &{\bfseries 0.7563} &{\bfseries 0.7627} &{\bfseries 0.7639} &{\bfseries 0.8564} &{\bfseries 0.8658} &{\bfseries 0.8697} &{\bfseries 0.8723} &{\bfseries 0.5925} &{\bfseries 0.6194} &{\bfseries 0.6330} &{\bfseries 0.6384}  \\
\hline
\end{tabular}
\label{table_MAP2}
\end{table*}

\begin{figure*}[ht]
\subfigure[NUS-WIDE]{
\begin{minipage}[h]{0.33\linewidth}
\centering
\includegraphics[width=1.\textwidth]{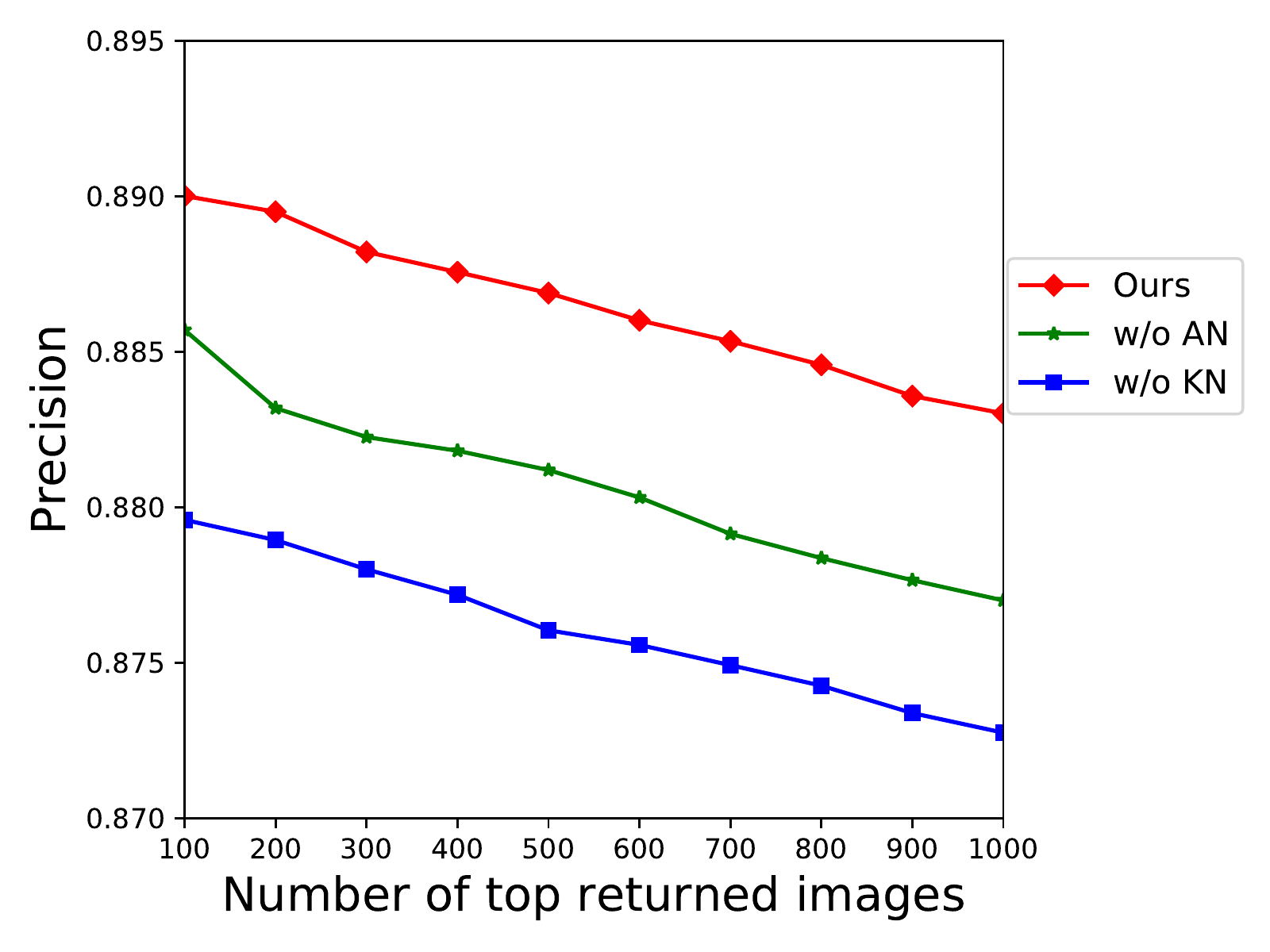}
\label{cifar:side:a}
\end{minipage}%
}
\subfigure[MIR-Flickr 25k]{
\begin{minipage}[h]{0.33\linewidth}
\centering
\includegraphics[width=1.\textwidth]{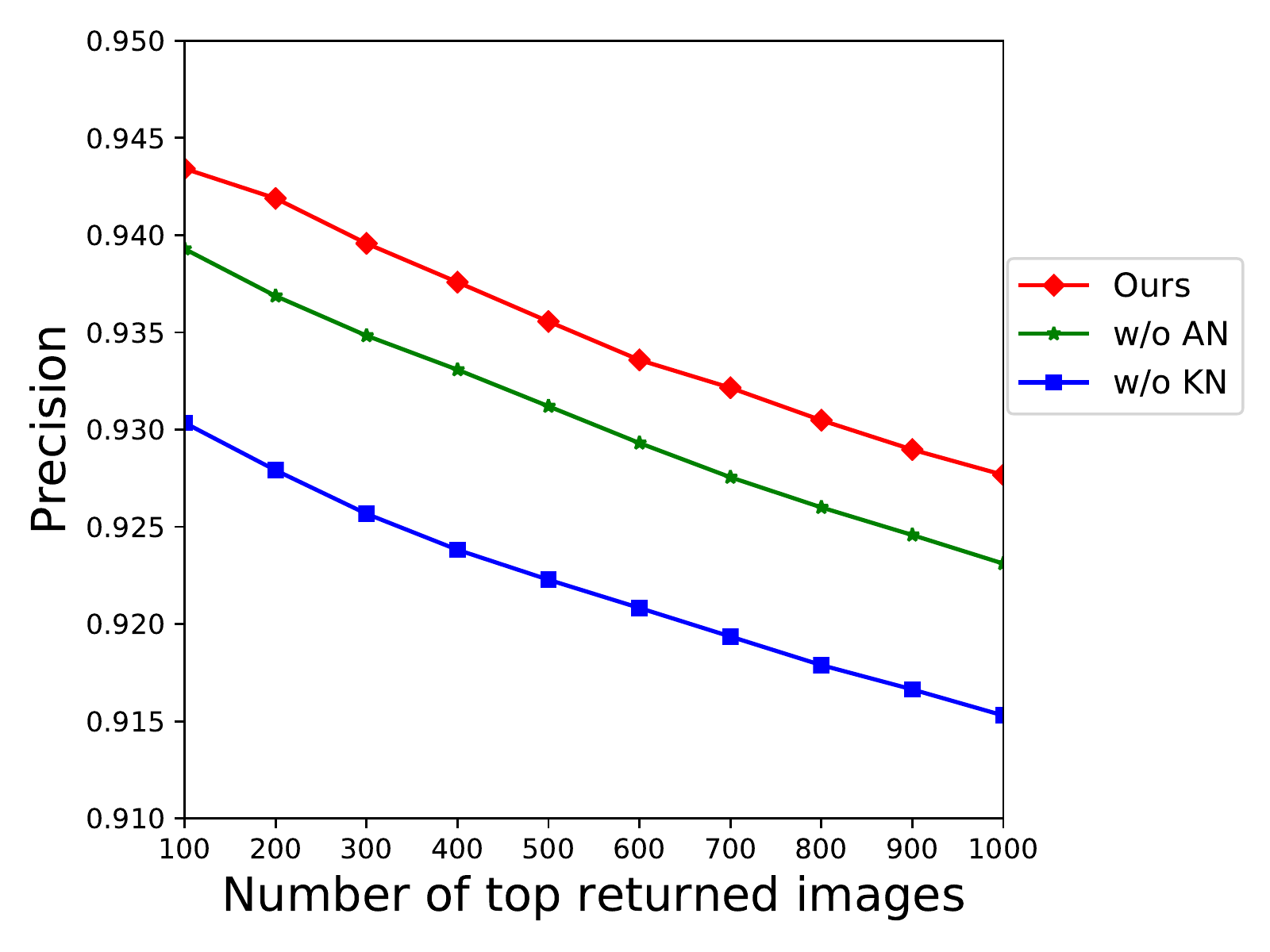}
\label{sun:side:b}
\end{minipage}
}
\subfigure[IAPR TC-12]{
\begin{minipage}[h]{0.33\linewidth}
\centering
\includegraphics[width=1.\textwidth]{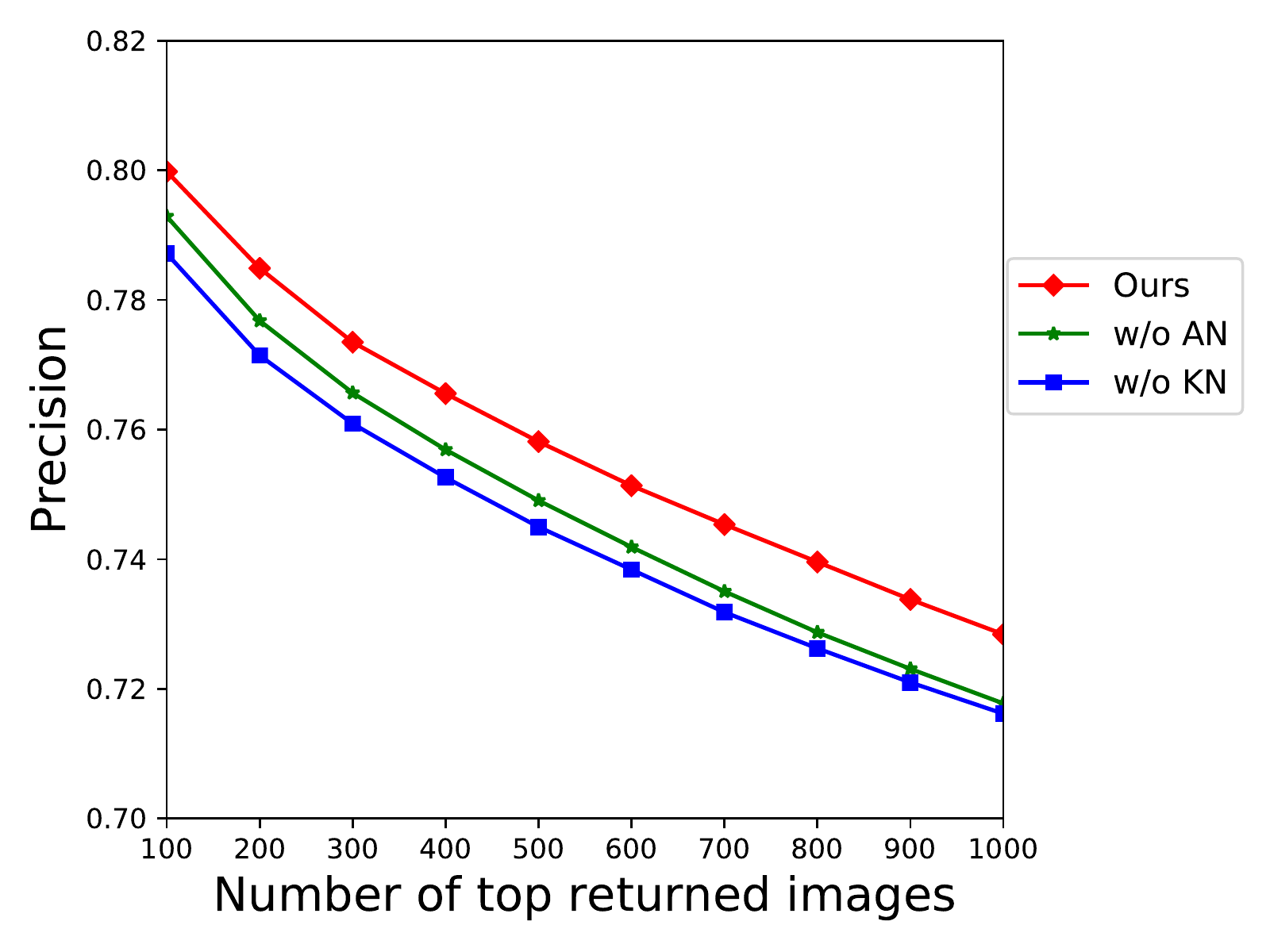}
\label{sun:side:b}
\end{minipage}
}
\caption {\small{The comparison results of precision curves for ablation study.}}
\label{Precision2}
\end{figure*}

\subsection{Comparison with State-of-the-art Methods}
In the first set of experiments, we compare the performance of the proposed method with state-of-the-art baselines. We evaluate two different approaches as the baselines.

The first set of baselines is the unimodal approaches. In this set of baselines, only one modality is used to train the hash functions. For the image modality, several state-of-the-art image hashing algorithms are selected: deep pairwise-supervised hashing (DPSH)~\cite{li_pairwise}, deep supervised hashing (DSH)~\cite{liu2016deep}, HashNet~\cite{cao2017hashnet} and deep triplet hashing (DTH)~\cite{lai2015simultaneous}. DPSH and DSH belong to deep pair-wise approaches, and DTH is a triplet-based approach. HashNet aims to minimize the quantization errors of the hash codes. For fair comparison, the deep architectures for these four methods are all the same as ours. For the text modality, we use the same network for text data, which is referred to as TextHash. TextHash only uses the text representations to learn the binary codes.

The second set of baselines is different fusion strategies used to combine multiple modalities. We note that only the fusion module in Figure~\ref{framework} uses different fusion strategies and the other modules are the same.
\begin{itemize}
\item \textbf{Concat} We concatenate both intermediate features of the image and text modalities to train the hashing architectures.

\item \textbf{GMU} A gate multimodal unit (GMU)~\cite{arevalo2017gated} is an internal unit in a neural network for data fusion. GMU uses multiplicative gates to determine how modalities influence the activation of the unit.

\item \textbf{MCB} Multimodal compact bilinear pooling (MCB)~\cite{fukui2016multimodal} uses bilinear pooling~\cite{lin2015bilinear} to combine visual and text representations.
\end{itemize}

Table~\ref{table_MAP} show the results of comparisons of the obtained MAP values for the three mulitmodal datasets. Figure~\ref{Precision-recall} and Figure~\ref{Precision} show the precision-recall and precision curves on 32 bits. Our proposed method yields the highest accuracy and beats all the baselines for most levels. Two observations can be made from the results as follows.

1) Compared with the unimodal approaches, our method performs significantly better than all baselines. For instance, our method yields the higher accuracy compared to the TextHash that only use the text modality. For image hashing methods, our method obtains a MAP of 0.7395 on 16 bits, compared with the value of 0.7115 of the HashNet on NUS-WIDE. On MIR-Flickr 25k, the MAP of DTH is 0.8332, while the proposed method is 0.8658 on 32 bits. The proposed method shows a relative increase of 4.6\%$\sim$6.9\% on the IAPR TC-12 compared to the DTH algorithm. Note that DTH and our method use the same triplet ranking loss function and DTH achieves an excellent performance. Even so, our method performances better than DTH. These results indicate that multi-modal approaches can improve the performance.

2) Compared with other deep fusion strategies, our method also yields the best performance on all databases. Firstly, compared to the \textit{Concat} approach, the only different is that using or not using the modal-aware operations, these comparisons can show us whether the modal-aware features can contribute to the accuracy or not.  The results indicate that our modal-aware features can achieve better performance. For example, the MAP of our proposed method is 0.7395 when the bit length is 16, compared to 0.7274 of \textit{Concat} on NUS-WIDE. Thus it is desirable to learn the powerful features for multi-modal retrieval. Compared to the GMU and MCB two baselines which achieve excellent performances, our proposed method also yields better performance.  The main reason is that our method can incorporate the information from other modalities to learn the intermediate features, while the intermediate features of GMU and MCB are learned via individual neural layers.

\subsection{Ablation Study}
In the second set of the experiments, an ablation study was perform to elucidate the impact of each part of our method on the final performance.

In the first baseline, we explore the effect of the kernel network. In this baseline, the attention network is fixed and we do not use the kernel network. That is the features are directly forwarded to the attention network and the only difference is that using or not using the kernel network, which is referred to as \textit{w/o KN}.

The second baseline explores the effect of the attention network. In this baseline, the kernel network was first performed to obtain two intermediate features. Then, we concatenate the two features to obtain the joint representation. We note that the only difference between the baseline and our method is the use or lack of use of the attention network. We use \textit{w/o AN} to denote the baseline that is not using the attention network.

The comparison results are shown in Table~\ref{table_MAP2} and Figure~\ref{Precision2}. The results show that our proposed method can achieve better performance than the two baselines. For instance, our method obtains a MAP of 0.7627 on 48 bits, compared to 0.7519 of the  \textit{w/o AN} and 0.7467 of the \textit{w/o KN}. The results indicate that it is desirable to learn the intermediate features with both the kernel network and the attention network.

In this paper, the text is represented as a bag-of-word vector. Other text representations, e.g., Sent2Vec or BERT~\cite{devlin2018bert}, can be used in our framework. For example, on IAPR TC-12 database, each image is associated with a text caption. Thus Sent2Vec, which is computed via the pre-trained model~\footnote{https://github.com/epfml/sent2vec}, can be used as the text representations. Table~\ref{table_MAP3} shows the comparison results with respect to MAP.

\begin{table}[ht]
\centering
\caption{The comparison results of different texts representation.}
\label{table1}
\begin{tabular}{|c|p{0.9cm}p{0.9cm}p{0.9cm}p{0.9cm}|}
\hline
\multirow{2}{*}{{Method}}&\multicolumn{4}{c|}{IAPR TC-12}\\
\cline{2-5}
&16bits & 32bits & 48bits & 64bits \\
\hline
BoW &0.5925 &0.6194 &0.6330 &{\bfseries 0.6384} \\
\hline
Sent2Vec &{\bfseries 0.5961} &{\bfseries 0.6232} &{\bfseries 0.6336} &0.6357 \\
\hline
\end{tabular}
\label{table_MAP3}
\end{table}

\section{Conclusion}
In this paper, we proposed a modal-aware operation for learning good feature representations. The key to success comes from designing a generic building block to capture the underlying correlation structures in heterogeneous multi-modal data prior to multimodal fusion. First, we proposed a kernel network to learn the non-linear relationships. The kernel similarities between two modalities were learned to reweight the original features. Then, we proposed an attention network, which aims to select the informative parts of the intermediate features. The experiments were conducted on three benchmark datasets, and the results demonstrate the appealing performance of the proposed modal-aware operations.


%





\ifCLASSOPTIONcaptionsoff
  \newpage
\fi



%
%
%

\bibliographystyle{IEEEtran}
\bibliography{egbib}

\end{document}